\documentclass[9pt,technote]{IEEEtran}

\usepackage{hyperref}
\usepackage{cite}
\usepackage{graphicx} % for pdf, bitmapped graphics files
\DeclareGraphicsExtensions{.pdf,.png,.jpg,.jpeg}	%imagenes
\usepackage{subcaption}
\usepackage{tabulary}
\usepackage{amsmath}

\DeclareMathOperator*{\argmin}{argmin}
\usepackage{amssymb}
\usepackage{gensymb}
\usepackage{xcolor}

\usepackage{array}
\newcolumntype{L}[1]{>{\raggedright\let\newline\\\arraybackslash\hspace{0pt}}m{#1}}
\newcolumntype{C}[1]{>{\centering\let\newline\\\arraybackslash\hspace{0pt}}m{#1}}
\newcolumntype{R}[1]{>{\raggedleft\let\newline\\\arraybackslash\hspace{0pt}}m{#1}}

\begin{document}

\title{Direct Sparse Mapping}

\author{Jon~Zubizarreta,
        Iker~Aguinaga,
        and~J. M. M.~Montiel% <-this % stops a space
\thanks{This work has been supported by the Basque Government under the predoctoral grant PRE\_2018\_2\_0035, under projects Langileok and Malgurob and by the Spanish Government under grant DPI2017-91104-EXP.}% <-this % stops a space
\thanks{J. Zubizarreta and I. Aguinaga are with the Computer Vision and Robotics Group at CEIT-BRTA, Spain {\tt\small jzgorostidi@gmail.com iaguinaga@ceit.es} J. M. M. Montiel is with I3A Universidad de Zaragoza, Spain {\tt\small josemari@unizar.es}}% <-this % stops a space
}

%\markboth{Short Paper. Journal of \LaTeX\ Class Files,~Vol.~14, No.~8, April~2020}


\onecolumn
\newpage
\thispagestyle{empty}

\begin{center}
\large

\vspace{7mm}

This paper has been accepted for publication in IEEE Transactions on Robotics.

\vspace{7mm}

DOI: \href{https://doi.org/10.1109/TRO.2020.2991614}{10.1109/TRO.2020.2991614}

IEEE Xplore: \href{https://ieeexplore.ieee.org/document/9102352}{https://ieeexplore.ieee.org/document/9102352}

\end{center}

\vspace{14mm}

\begin{large}
\copyright2020 IEEE. Personal use of this material is permitted.  Permission from IEEE must be obtained for all other uses, in any current or future media, including reprinting/republishing this material for advertising or promotional purposes, creating new collective works, for resale or redistribution to servers or lists, or reuse of any copyrighted component of this work in other works.
\end{large}

\twocolumn
\maketitle
\pagenumbering{arabic} 

\begin{abstract}
Photometric bundle adjustment, PBA, accurately estimates geometry from video. However, current PBA systems have a temporary map that cannot manage scene reobservations. We present, DSM, a full monocular visual SLAM based on PBA. Its persistent map handles reobservations, yielding the most accurate results up to date on EuRoC for a direct method.
\end{abstract}


% \fi
\IEEEpeerreviewmaketitle

\section{Introduction}

% Photometric Bundle Adjustment
\IEEEPARstart{P}{hotometric} bundle adjustment has proven to be an effective method for estimating scene geometry and camera motion in Visual Odometry (VO) \cite{Engel2016a}. As a direct optimization, PBA minimizes the photometric error of map point observations over a local sliding-window of active keyframes. The number of active keyframes is limited to avoid large computations. Points are sampled across image pixels with locally high gradient module, such as edges and weak intensity variations. They are associated to only one keyframe where they are initialized. In the rest of keyframes, there is not an explicit and fixed data association, because the PBA recomputes the correspondences as a part of the optimization. Thus, direct methods do not rely on the repeatability of selected points and are able to operate in scenes with low texture but with contours.

% Visual Odometry (VO)
Current PBA based methods are only able to do VO, which builds a temporary map to precisely estimate the camera pose. They use a sliding-window that selects close in time active keyframes, marginalizing map points that leave the field of view. This strategy reduces the computation complexity by removing old cameras and points while maintaining the system consistent to unobservable degrees of freedom, i.e. absolute pose and scale. Hence, if the camera revisits already mapped areas, the PBA cannot reuse marginalized map points and it is forced to duplicate them. This is a severe limitation: the system cannot benefit from the highly informative reobservations of map points, and this causes motion drift and structure inconsistencies.

% Visual SLAM
In contrast, VSLAM methods build a persistent map of the scene, and continuously process map point reobservations. Instead of using a sliding-window and marginalization, they retain keyframes and map points with a fixed location in the model and select the active keyframes and map points according to covisibility criteria, i.e. they observe several map points in common. This results in a network of keyframes where the connectivity is based on whether they observe the same scene region even if they are far in time. The fixation strategy maintains the system consistent to unobservable degrees of freedom (gauge freedom) and it enables the reuse of map points. Thus, VSLAM approaches can extract the rich information of map point reobservations reducing the drift in the estimates.

\begin{figure}
	\centering
	\includegraphics[width=0.45\textwidth]{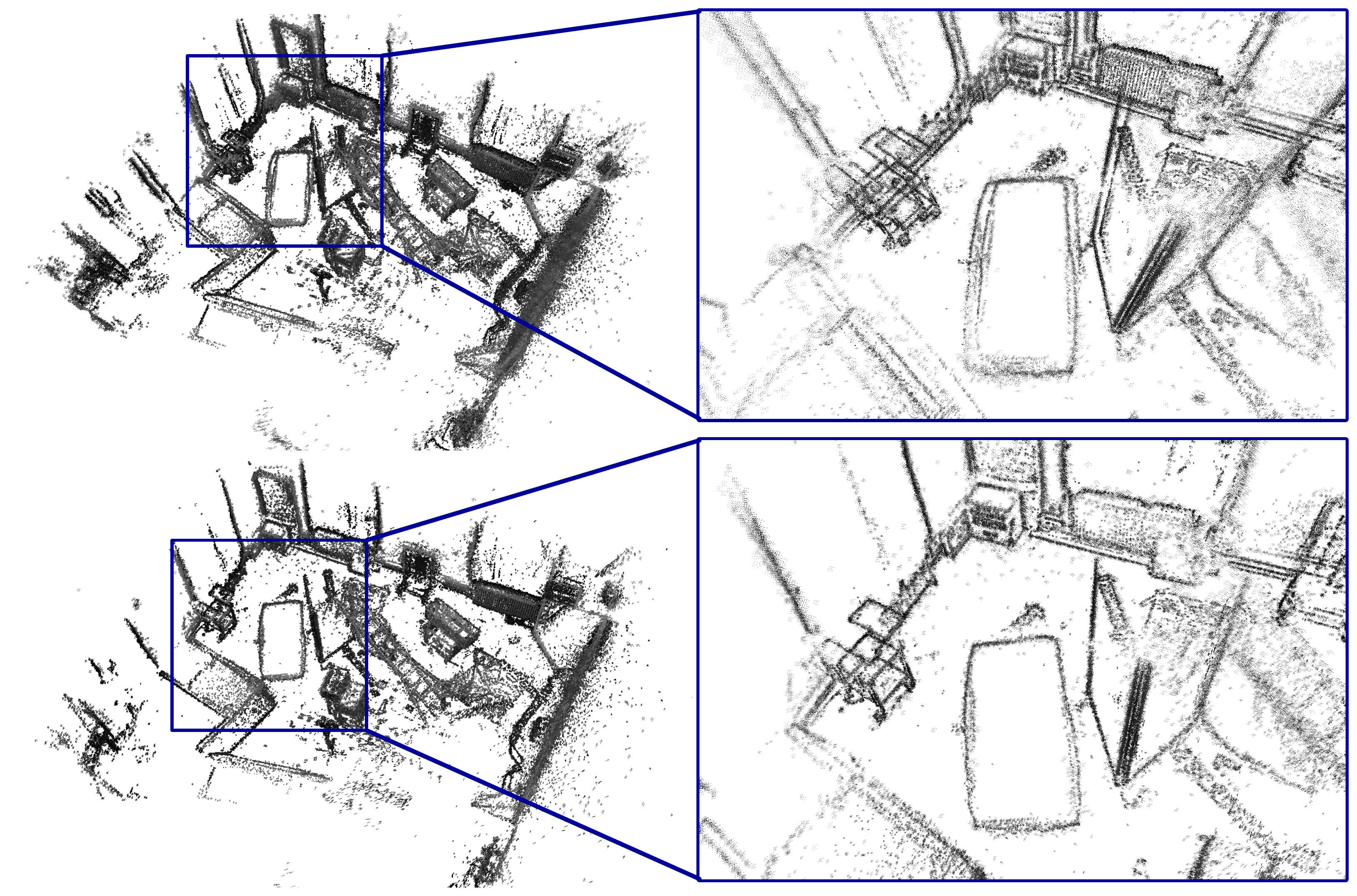}
	\caption{Estimated map by DSM with (bottom) and without (top) point reobservations in the V2\_01\_easy sequence of the EuRoC MAV dataset. DSM can produce consistent maps without duplicates.}
	\label{fig:example_V21}
\end{figure}

% Challenges of PBA with VM scheme
Transforming PBA based direct VO systems into VSLAM  is not straightforward because there are several challenges to solve. First, when the camera revisits already mapped areas, the system has to select active keyframes that include map point reobservations. They are difficult to obtain because there are not point correspondences between keyframes. At the same time, we have to guarantee accurate map expansion during exploration. We propose to select the active keyframes according to a combination of temporal and covisibility criteria. In this way, the PBA includes in the optimization keyframes that observe the active scene region with high parallax even if they are far in time. Second, the PBA includes map points and keyframes distant in time and, hence, affected by the estimation drift. Normally, the photometric convergence radius is around 1-2 pixels due to image linearization and, thus, a standard PBA cannot compensate the drift. We propose a multiscale PBA optimization to handle successfully these convergence difficulties. Third, we have to ensure the robustness of the PBA against spurious observations. They mainly arise from the widely separated active keyframes -- in contrast to the close keyframes of VO -- which render occlusions and scene reflections that violate the photo-consistency assumption. We incorporate a robust influence function based on the t-distribution into the PBA to handle the adverse effect of these observations.

% contributions
We present a new direct VSLAM system, DSM (Direct Sparse Mapping). Up to our knowledge, this is the first fully direct monocular VSLAM method that is able not only to detect point reobservations but also to extract the rich information they provide (see Fig.\,\ref{fig:example_V21}). In summary, we make the following contributions:

\begin{itemize}
	
	% map
	\item A  persistent map which allows to reuse existing map information directly with the photometric formulation.
	
	% local window
	\item The  Local Map Covisibility Window (LMCW)  criteria to select the active keyframes that observe the same scene region, even if they are not close in time, and the map point reobservations.
	
	% multi scale photometric BA
	\item We show that the PBA needs a coarse-to-fine scheme to convergence. This exploits the rich geometrical information provided by point reobservations from keyframes rendering high parallax.
	
	% error distribution
	\item We show that a t-distribution based robust influence function together with a pixel-wise outlier management strategy increases the PBA consistency against outliers derived from the activation of distant keyframes.
	
	% experimental validation
	\item An experimental validation of DSM in the EuRoC dataset \cite{Burri2015}. We report quantitative results of the camera trajectory and, for the first time, of the reconstructed map. We obtain the most accurate results among direct monocular methods so far.
	
	% open-source
	\item We make our implementation publicly available\footnote{https://github.com/jzubizarreta/dsm}.
	
\end{itemize}

\section{Related Work}

% indirect VSLAM methods
The first real-time monocular VSLAM methods were indirect approaches, using FAST and Harris corners associated across images in the form of 2D fixed correspondences. The 3D geometry was estimated minimizing the reprojection error. They relied on the repeatability of the corner detectors and required rich visual texture. Thank to the feature descriptors they associate distant images. Davison et al. present MonoSLAM \cite{Davison2007}, which recovers the scene geometry in an EKF-based framework, later extended in \cite{Civera2008} to include a parametrization in inverse depth. Klein and Murray in PTAM \cite{Klein2007} propose for the first time to parallelize the tracking and mapping tasks, demonstrating the viability of using a BA scheme to maintain a persistent map in small workspaces. Later, \cite{Strasdat2011} proposes a double window optimization to extend the potential of feature-based VSLAM to long-term applications. It combines a local BA with a global pose-graph optimization using covisibility constraints based on point matches. Following these works, ORB-SLAM \cite{Mur-Artal2015} presents which is the reference solution among indirect VSLAM approaches. Up to date, it is the most accurate monocular VSLAM method in many scenarios. The key aspect of its precision comes from the management of map point reobservations in the BA using an appearance based covisibility graph. Similarly, DSM transfers the main ideas of indirect VSLAM techniques, to direct systems significantly increasing the accuracy of their estimates. As a direct approach, DSM does not compute explicit point matches and, thus, cannot build an appearance based covisibility graph. Instead, DSM relies on geometric constraints to build covisibility connections between far in time keyframes. In addition, it works with a smaller window of covisible keyframes than ORB-SLAM to control the computational limitations.

% vo with sliding-window + marginalization
Recently, VO approaches have shown impressive performance. SVO \cite{Forster2014} proposes an hybrid approach to build a semi-direct VO system. It uses direct techniques to track and triangulate points but ultimately optimizes the reprojection error of those points in the background. OKVIS \cite{Leutenegger2015} presents a feature-based visual-inertial odometry system that continuously optimizes the geometry of a local map marginalizing the rest. Recently, Engel et al. \cite{Engel2016a} made a breakthrough with their DSO, the first fully direct VO approach that jointly optimizes motion and structure formulating a PBA and including a photometric calibration into the model. Inspired by OKVIS, DSO performs the optimization over a sliding-window, where old keyframes as well as points that leave the field of view of the camera are marginalized. It has shown impressive odometry performance and it is the reference among direct VO methods. However, as a pure VO approach DSO cannot reuse map points once they are marginalized which causes camera localization drift and map inconsistencies. DSM uses the same photometric model of DSO and goes one step further to build the first fully direct VSLAM solution with a persistent map.

% vo + pose-graph
Many VO systems have been extended to cope with loop closures. Most propose to include a feature-based Bag of Binary Words (DBoW) to detect loop closures and estimate pose constraints between keyframes, following \cite{Galvez-Lopez2012}. Then, a pose-graph optimization finds a correction for the keyframe trayectory. VINS-mono \cite{Qin2018} uses a similar front-end to OKVIS but includes additional BRIEF features to perform loop closure. LSD-SLAM \cite{Engel2014} was the first direct monocular VO for large-scale environments. The method recovers semi-dense depth maps using small-baseline stereo comparisons and reduces accumulated drift with a pose-graph optimization. Loop closures are detected using FAB-MAP \cite{Cummins2008}, an appearance loop detection algorithm, which uses different features to those of the direct odometry. LDSO \cite{Gao2018} extends DSO with a conventional ORB-DBoW to detect loop closures and reduce the trajectory drift by pose-graph optimization. All these methods have the next drawbacks: (1) they uses a different objective function and points to those of the odometry; (2) loop closure detection relies on feature repeatability, missing many corrections; (3) the error correction is distributed equally over keyframes, which may not be the optimal solution; (4) although the trajectory is spatially corrected, existing information from map points is not reused and, thus, ignored during the optimization. In contrast, full VSLAM systems like ORB-SLAM and DSM reuse the map information thanks to its persistent map. The reobservations are processed with their standard BA (either geometric or photometric), resulting in more accurate estimates. Thanks to the improvement in accuracy the need of loop closure detection and correction is postponed to trajectories longer than in their VO counterparts.

% t-distribution
Moreover, DVO \cite{Kerl2013} proposes a probabilistic formulation for direct image alignment. Inspired by \cite{Lange1989}, they show the robustness of using a t-distribution to manage the influence of noise and outliers. \cite{Babu2016} demonstrates that the t-distribution represents well photometric errors while not geometric errors. We incorporate these ideas into the sparse photometric model together with a novel outlier management strategy. In this way, we make the non-linear PBA optimization robust to spurious point observations. They normally appear as a result of widely separated active keyframes and lack of explicit point matches.

\section{Direct Mapping}

% parallel tracking and mapping
The proposed VSLAM system consists of a tracking front-end (Sec. \ref{sec:frontend}) and an optimization back-end (Sec. \ref{sec:PBA}). The front-end tracks frames and points, and also provides the coarse initialization for the PBA. The back-end determines which keyframes form the local window (Sec. \ref{sec:lmcw}) and jointly optimizes all the active keyframes and map point parameters. Similarly to most VSLAM systems \cite{Klein2007,Mur-Artal2015,Engel2016a}, the front-end and the back-end run in two parallel threads:

\begin{enumerate}
	\item The tracking thread obtains the camera pose at frame rate. It also decides when the map needs to grow by marking some of the tracked frames as keyframes.
	\item The mapping thread processes all new frames to track points from active keyframes. Besides, if the new frame is marked as a keyframe, the local window is recalculated, new points are activated and the PBA optimizes motion (keyframes) and structure (points) together using active keyframes. Finally, it maintains the model globally consistent, i.e. removes outliers, detects occlusions and avoids point duplications (Sec. \ref{sec:outlier}).
\end{enumerate}

% mapping structure
The persistent map is composed of keyfames that are activated or deactivated according to covisibility criteria with the latest keyframe. The absolute pose of a keyframe $i$ is represented by the transformation matrix $\mathbf{T}_{i} \in SE(3)$. For each keyframe, we select as candidate points those with a locally high gradient module and spread over the image. Each map point $\mathbf{p}$ is created in a keyframe (Sec. \ref{sec:frontend}) and its pose is coded as its inverse depth $\rho = p_z^{-1}$. Thus, for each keyframe we store the raw image and the associated map points. We assume all images to be undistorted. We use the pinhole model to project a point from 3D space to the image plane, $\mathbf{u} = \pi(\mathbf{p}) = \mathbf{K} (p_x/p_z, p_y/p_z, 1)^T$, where $\mathbf{K}$ is the camera matrix. Its inverse is also defined when the inverse depth of the point is known $\mathbf{p} = \pi^{-1}(\mathbf{u}, \rho) = \rho^{-1} \mathbf{K^{-1}} (u_x, u_y, 1)^T$.

The LMCW (Sec. \ref{sec:lmcw}) selects which keyframes are active and form the local window. Once a keyframe is active, all its parameters (pose and affine light model) and associated points (inverse depth) are optimized by the PBA. Otherwise, they remain fixed to maintain the system consistent to unobservable degrees of freedom. During optimization, we will use $\boldsymbol\xi \in SE(3)^n \times \mathbb{R}^{2n+m} $ to represent the set of optimized parameters ($n$ keyframes and $m$ points) and $\delta\boldsymbol\xi \in se(3)^n \times \mathbb{R}^{2n+m}$ to denote the increments. Moreover, we use the left-compositional convention for all optimization increments, i.e. $\boldsymbol\xi^{(t+1)} = \delta\boldsymbol\xi^{(t)} \boxplus \boldsymbol\xi^{(t)}$. This direct VSLAM framework enables to build a persistent map and reuse existing map information from old keyframes directly in the photometric bundle adjustment.

\subsection{Photometric Model} \label{sec:model}
The same photometric function, the one proposed in  \cite{Engel2016a},  is used in the whole system, i.e. geometry initialization (camera and point tracking), local windowed PBA and map reuse. For each point $\mathbf{p}$, we evaluate the sum of square intensity differences $r_k$ over a small patch $\mathcal{N}_p$ around it between the host $I_{i}$ and target $I_j$ images. We include an affine brightness transfer model to handle the camera automatic gain control and changes in scene illumination. The observation of a point $\mathbf{p}$ in the keyframe $I_j$ is coded by:
\begin{equation}
E_p = \sum_{\mathbf{u}_k \in \mathcal{N}_p} w_k \bigg((I_{i}[\mathbf{u}_k] - b_{i}) - \frac{e^{a_i}}{e^{a_j}}(I_j[\mathbf{u'}_k] - b_j)\bigg)^2,
\label{eq:model}
\end{equation}
\noindent
where $\mathbf{u}_k$ is each of the pixels in the patch; $\mathbf{u'}_k$ is the projection of $\mathbf{u}_k$ in the target frame with its inverse depth $\rho_k$, given by $\mathbf{u'}_k = \pi(\mathbf{T}_{j,i}\cdot \pi^{-1}(\mathbf{u}_k, \rho_k))$ with $\mathbf{T}_{j,i}=\mathbf{T}_{j}^{-1}\mathbf{T}_{i}$; $a_{i}, b_{i}, a_{j}, b_{j}$ the affine brightness functions for each frame; and $w_k = w_{r_k} w_{g_k}$ a combination of the robust influence function $ w_{r_k}$ and a gradient dependent weight $w_{g_k}$:
\begin{equation}
w_{g_k} = \frac{c^2}{c^2 + \parallel \nabla I\parallel_2^2},
\label{eq:weight}
\end{equation} 
\noindent
which works as a heuristic covariance in the Maximum Likelihood (ML) estimation, reducing the influence of high gradient pixels due to noise. To sum up, the photometric cost function (\ref{eq:model}) depends on geometric ($\mathbf{T}_i, \mathbf{T}_j, \rho$) and photometric parameters ($a_i, b_i, a_j, b_j$).

\subsection{Photometric Bundle Adjustment (PBA)} \label{sec:PBA}
Every time a new keyframe is created, all model parameters are optimized by minimizing the error from Eq. (\ref{eq:model}) over the LMCW of active keyframes $\mathcal{K}$. The total error is given by:
\begin{equation}
E = \sum_{I_i \in \mathcal{K}} \sum_{\mathbf{p} \in \mathcal{P}_i} \sum_{j \in obs(\mathbf{p})} \sum_{\mathbf{u}_k \in \mathcal{N}_p} w_k r^2_k(\boldsymbol\xi),
\label{eq:fullModel}
\end{equation}
where $\mathcal{P}_i$ is the set of points in $I_i$ and $obs(\mathbf{p})$ the set of observations for $\mathbf{p}$. Note that the LMCW reuses map point observations for which the initial solution is not inside the convergence radius and, thus, the PBA is not able to correct. Hence, we propose to use a coarse-to-fine optimization scheme over all active keyframes. In each level, we iterate until convergence and use the estimated geometry as an initialization for the next level. The same points are used across all levels and each level is treated independently, i.e. neither the influence function nor outlier decisions are propagated across the levels (Sec. \ref{sec:robustPBA}). In this way, we are able to handle larger camera and point increments $\delta\boldsymbol\xi$ with the photometric model.

We minimize Eq. (\ref{eq:fullModel}) using the iteratively re-weighted Levenberg-Marquardt algorithm. From an initial estimate $\boldsymbol\xi^{(0)}$, each iteration $t$ computes weights $w_k$ and photometric errors $r_k$ to estimate an increment $\delta\boldsymbol\xi^{(t)}$ by solving for the minimum of a second order approximation of Eq. (\ref{eq:fullModel}), with fixed weights:
\begin{equation}
\delta\boldsymbol\xi^{(t)} = -\mathbf{H}^{-1}\mathbf{b},
\end{equation}
with $\mathbf{H} = \mathbf{J}^T\mathbf{W}\mathbf{J} + \lambda \textnormal{diag}(\mathbf{J}^T\mathbf{W}\mathbf{J})$, $\mathbf{b} = \mathbf{J}^T\mathbf{W}\mathbf{r}$ and $\mathbf{W} \in \mathbb{R}^{m\times m}$ is a diagonal matrix with the weights $w_k$, $\mathbf{r}$ is the error vector and $\mathbf{J} \in \mathbb{R}^{m\times d}$ is the Jacobian of the error vector with respect to a left-composed increment given by:
\begin{equation}
\mathbf{J}_k = \frac{\partial r_k (\delta\boldsymbol\xi \boxplus \boldsymbol\xi^{(t)})}{\partial \delta\boldsymbol\xi} \biggr\rvert_{\substack{\delta\boldsymbol\xi = 0}}.
\end{equation}

The PBA is implemented using Ceres optimization library \cite{Agarwal} with analytic derivatives. Image gradients are computed using central pixel differences at integer values. For subpixel intensity and gradient evaluation, bilinear interpolation is applied. We take advantage of the so-called primary structure and use the Schur complement trick to solve the reduced problem \cite{Triggs2000}. The gauge freedoms are controlled fixing all other keyframes that are covisible with the active ones.

% -----------------------------------------------------------------------------------
\section{LMCW: Local Map Covisibility Window} \label{sec:lmcw}
\begin{figure}
    \centering
    \includegraphics[width=0.42\textwidth]{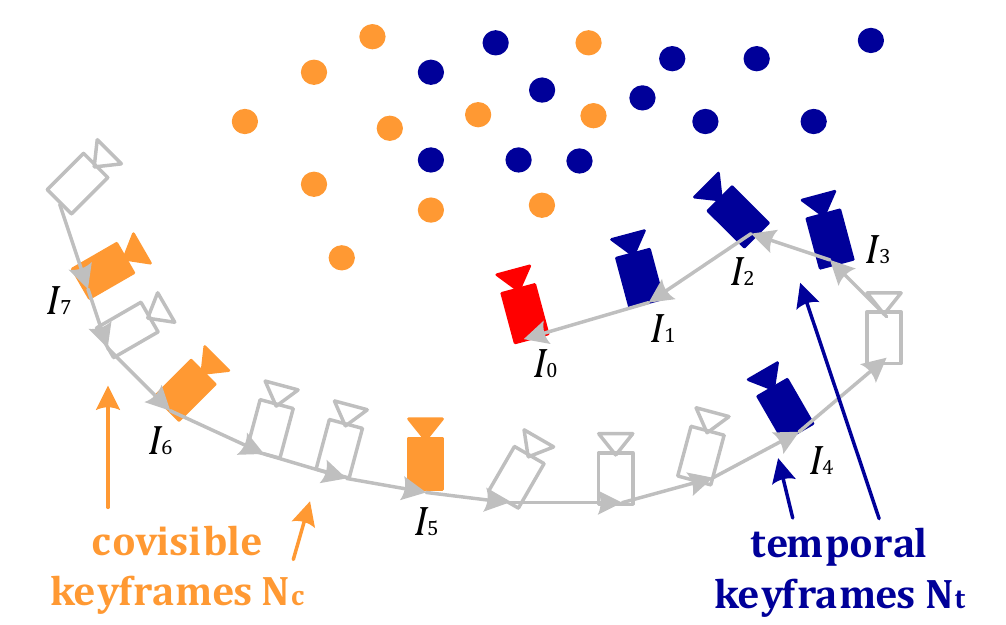}
    \caption{LMCW example with $N_w=7$ and the latest keyframe being created (red). It is composed of $N_t=4$ temporal (blue) and $N_c=3$ covisible (orange) active keyframes.}
    \label{fig:LMCW_example}
\end{figure}

This section presents the LMCW and the strategy to select its active keyframes and active map points. It is a combination of temporal and covisible criteria with respect to the latest keyframe being created. The LMCW is composed of two main parts: the temporal and the covisible. Fig. \ref{fig:LMCW_example} shows the LMCW selection strategy.

The first part is composed of $N_t$ temporally connected keyframes that form a sliding-window like in \cite{Engel2016a}. This part is critical during exploration because it initializes new points (Sec. \ref{sec:frontend}) and maintains the accuracy in odometry. Whenever a new keyframe is created, we insert it into the temporal part and remove another one. Thus, we maintain fixed size temporal keyframes. The strategy that selects the removed keyframe from the temporal part is summarized as:

\begin{enumerate}
    \item Keep the last two keyframes ($I_1$ and $I_2$) to ensure the odometry accuracy during challenging exploratory motions, such as rotations. It avoids premature fixation of keyframes location, guaranteeing that keyframes are well optimized beforehand.  
    \item The remaining keyframes are evenly distributed in space. We drop the keyframe $I_i$ that maximizes:    
    \begin{equation}
        s(I_i) = \sqrt{d(I_0, I_i)} \quad \textstyle{\sum}_{j=1}^{N_t}  \left(d\left(I_i,I_j\right)\right)^{-1},
    \end{equation}
	where $d(I_i,I_j)$ is the $L_2$ distance between keyframes $I_i$ and $I_j$. This strategy favors observations rendering high parallax into the PBA, which increases the accuracy.
\end{enumerate}
 
The second part is composed of $N_c$ covisible keyframes with those in the temporal part. Additionally, we seek to fill the latest keyframe $I_0$ with reobserved map points, favoring map points imaged in depleted areas (image areas where no other map points are imaged). Our strategy to achieve this goal is summarized as:

\begin{enumerate}
    \item Compute the distance map to identify the depleted areas.  All the map points from the temporal part are projected into the latest keyframe, then the distance map registers, for every pixel, the $L_2$ distance to its closest map point projection.
    \item Select a keyframe, among the list of old keyframes, that maximizes the number of projected points in the depleted areas using the distance map. We discard points that form a viewing angle above a threshold to detect and remove potential occluded points as early as possible.
    \item Update the distance map to identify the new depleted areas.
    \item Iterate from (2) until $N_c$ covisible keyframes are selected or no more suitable keyframes are found.
\end{enumerate}

The covisible part incorporates already mapped areas in the LMCW before activating new map points. The proposed strategy avoids map point duplications ensuring the map consistency. The values of $N_t$ and $N_c$ are tuned experimentally in Sec. \ref{sec:results}.

\section{Robust Non-linear PBA} \label{sec:robustPBA}
The LMCW selects widely separated active keyframes according to geometric criteria without any consideration about the actual photo-consistency between the images of the map points in the selected keyframes. Hence, it is possible that some of the points do not render photo-consistent images, because they suffer, for example, from occlusions or scene reflections.  

To make our PBA robust with respect to this lack of photo-consitency, we propose an outliers management strategy based on the photometric error distribution, from which we derive the appropriate weights for Eq. \ref{eq:fullModel}. According to the probabilistic approach, optimizing  the Eq. \ref{eq:fullModel} is equivalent to minimizing the negative log-likelihood of model parameters $\boldsymbol\xi$ given independent and equally distributed errors $r_k$,
\begin{equation} \label{eq:MAP}
\boldsymbol{\xi^*} = \argmin_{\boldsymbol\xi} - \sum_{k}^{n} \log p(r_k \mid \boldsymbol\xi) 
\end{equation}

The minimum of Eq. \ref{eq:MAP} is computed equating their derivatives to zero. This is equivalent to minimizing the re-weighted least-squares Eq. \ref{eq:fullModel} with the following weights:
\begin{equation} \label{eq:Weights}
w(r_k) = - \frac{\partial \log p(r_k)}{\partial r_k} \frac{1}{r_k}
\end{equation}

Therefore, the solution is directly affected by the photometric error distribution $p(r_k)$ (see \cite{Kerl2013} for further details). Next we consider different distributions.

\paragraph{Gaussian distribution} If errors are assumed to be normally distributed around zero $\mathcal{N}(0,\sigma_n^2)$, the model of error distribution is $p(r_k) \propto \text{exp}(r_k^2/\sigma_n^2)$. This model leads to a constant distribution of weights which is a standard least squares minimization. Thus, it treats all points equally and outliers cannot be neutralized: 
\begin{equation} \label{eq:normal_w}
w_n(r_k) = \frac{1}{\sigma_n^2}
\end{equation}

\paragraph{Student's t-distribution} Recently, \cite{Kerl2013} has analyzed the distribution of dense photometric errors for RGB-D odometry. It showed that the t-distribution explains dense photometric errors better than a normal distribution, providing a suitable weight function:
\begin{equation} \label{eq:t_dist_w}
w_t(r_k) = \frac{\nu + 1}{\nu + (\frac{r_k}{\sigma_t})^2}, \quad \text{when} \; \mu=0
\end{equation}

\begin{figure}
    \centering
    \includegraphics[width=0.4\textwidth]{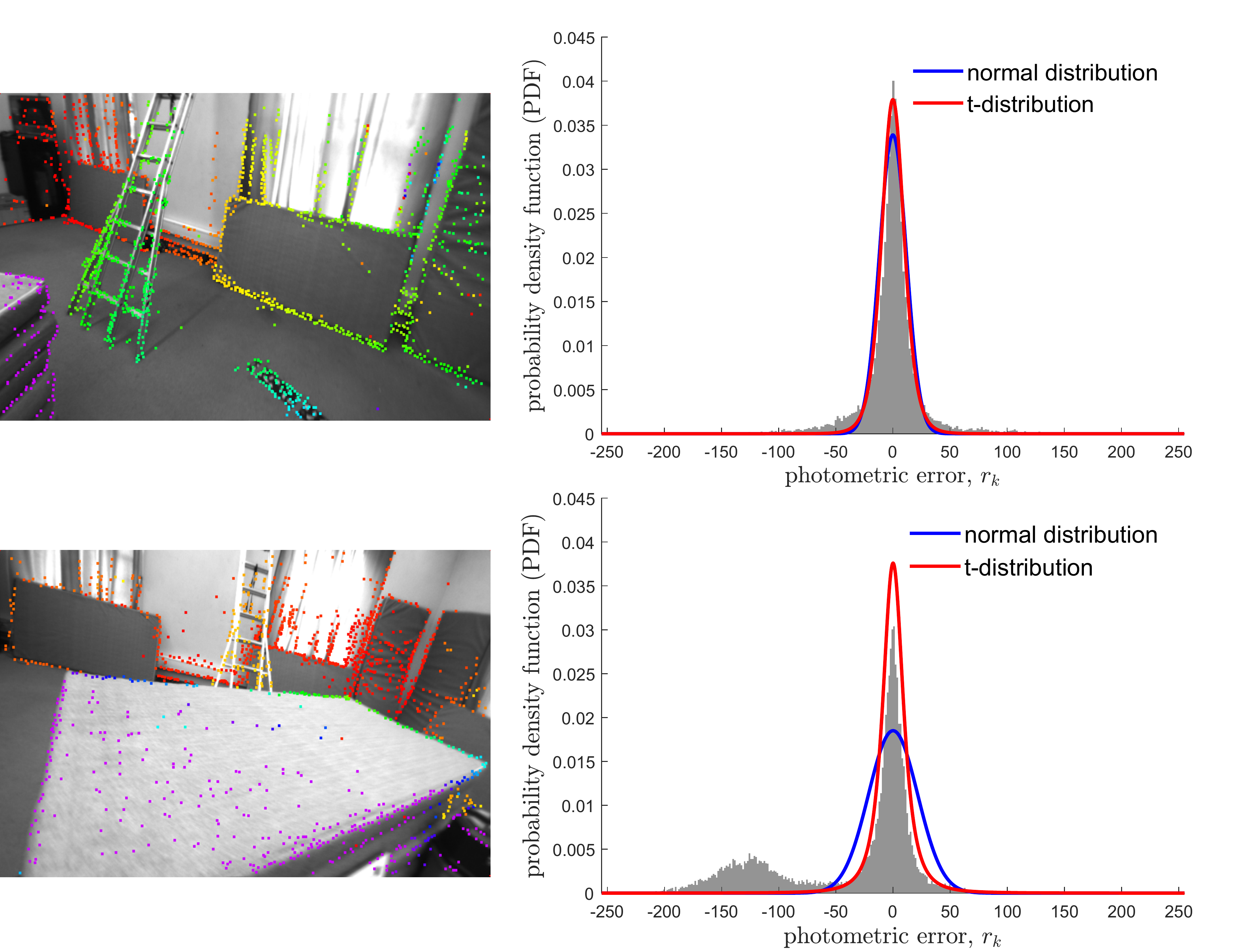}
    \caption{Probabilistic error modeling. The top row shows the case where most of the map points are photo-consistent, then both normal and t-distribution models fit well the photometric errors. The bottom row shows a challenging situation where covisible reobservations introduce many outliers due to occlusions, the t-distribution fits the observed errors better than the normal. On the left, the keyframe along with the point depth map after outlier removal.}
    \label{fig:dist_example}
\end{figure}

We have experimentally studied the sparse photometric errors and we conclude that the t-distribution also explains the sparse model properly (Fig. \ref{fig:dist_example}). In contrast to the normal distribution, the t-distribution quickly drops the weights as errors move to the tails, assigning a lower weight to outliers. Besides, instead of fixing the value of the degrees of freedom $\nu=5$ as in \cite{Kerl2013}, we study the behavior of the model when $\nu$ is fitted together with the scale $\sigma_t$ (see Sec. \ref{sec:results}). To fit the t-distribution, we minimize the negative log-likelihood of the probability density function with respect to $\nu$ and $\sigma_t$ using the gradient free iterative Nelder-Mead method \cite{LagariasJeffrey1998}. Besides, we filter out the gross outliers before fitting the t-distribution. We approximate the scale value $\hat{\sigma}$ using the Median Absolute Deviation (MAD) as $\hat{\sigma}=1.4826 \text{ MAD}$ and reject errors that $r_k > 3 \hat{\sigma}$.

\paragraph{M-estimators - Huber} Whether the distribution of errors is hard to know or it is assumed to be normally distributed, using M-estimators is a popular solution. One of the most popular ones is the Huber estimator, since it does not totally remove high error measurements but it decreases their influence, which is crucial for reobservation processing. The Huber weighting function is defined as:
\begin{equation} \label{eq:huber}
w_h(r_k) = 
\begin{cases}
\frac{1}{\sigma_n^2} & \text{if } |r_k| < \lambda \\
\frac{\lambda}{\sigma_n^2|r_k|} & \text{otherwise}
\end{cases}
\end{equation}
where $\lambda$ is usually fixed or dynamically changed each time step with the value $\lambda=1.345\sigma_n$ for $\mathcal{N}(0,\sigma_n^2)$. In this case, Huber gives linear influence to the outliers.

\subsection{Implementation of the probabilistic model into the PBA}
We have studied the error distribution in each keyframe and concluded that there are differences between them. These variations might come from motion blur, occlusions or noise (see Fig. \ref{fig:dist_example} and the accompanying video). Hence, we fit the error distribution for each keyframe separately using all the observations from active points in that keyframe. This allows to adapt the PBA to different situations, e.g. certain error values might be considered as an outlier in a regular keyframe but inlier in a challenging one due to motion blur.

Computing the error distribution and, thus, the weight distribution each iteration changes the objective function (Eq. \ref{eq:MAP}) and the performance of the optimization might degrade. We propose to compute the error distribution only at the beginning of each pyramid level and maintain it fixed during all the optimization steps. At the end of the PBA, the error distribution is recomputed again using the photometric errors obtained from the best geometry solution $\boldsymbol{\xi^*}$.

\subsection{Outlier management} \label{sec:outlier}
It is crucial to detect and remove outlier observations as soon as possible to maintain the PBA stability. To achieve this, we exploit the information from each observation, which includes measurements from eight different pixels. We propose to build a mask for each point and mark each pixel measurement $r_k$ as inlier or outlier. This helps handling points in depth discontinuities where other SLAM approaches typically struggle. To consider a pixel measurement as inlier, the photometric error has to be lower than the 95\% percentile of the error distribution of the target keyframe. For challenging keyframes the threshold will be higher, being more permissive, whereas for regular ones it will be lower, being more restrictive. When the current local PBA is finished, we count the number of inlier pixels in the mask. Whenever an observation contains a number of outlier pixels larger than a 30\%, the observation is marked as an outlier and removed from the list of observations of the point. Besides, during the optimization, if the number of outlier pixels is larger than a 60\%, the observation is directly discarded from the current optimization step, i.e. $w(r)=0$.

We also detect and remove outlier points from the map. We propose to control the number of observations in each point to decide if it is retained. To retain a new point, it must be observed in all the new keyframes after its creation, when it has been observed in three keyframes it is considered a mature point. Mature points are removed if the number of observations falls below three.

\section{Front-End} \label{sec:frontend}

\paragraph{Frame Tracking}
Each new frame is tracked against a local map, which is updated after every new keyframe decision. The local map is formed with active points from the LMCW referenced to the latest keyframe. The frame pose and its affine brightness transfer model are computed by minimizing Eq. \ref{eq:model} in which the map points and the latest keyframe remain fixed. The initial estimation is provided by a velocity model. We  use a coarse-to-fine optimization, as proposed in the PBA, to handle initial guesses with large errors.  We use the same robust influence function of Sec. \ref{sec:robustPBA} to reduce the impact of high photometric errors. In addition, we use the inverse compositional approach \cite{Baker2004} to avoid re-evaluating Jacobians each iteration and reduce the computational cost.

\paragraph{New Keyframe Decision}
Whenever we move towards unexplored areas, the map is expanded with a new keyframe. We use three different criteria with respect to the latest keyframe to decide if the tracked frame becomes a  keyframe:

\begin{enumerate}
	% point visibility
	\item The map point visibility ratio between the latest keyframe and the tracked frame, i.e. $s_u = N^{-1}\sum \textnormal{min}(p_z/p'_z, 1)$, where $N$ is the total number of visible points in the latest keyframe, $p_z$ the point inverse depth in the latest keyframe and $p'_z$ the point inverse depth in the tracked frame. The score is formulated to create more keyframes if the camera moves closer.
	% tracking parallax
	\item The tracked frame parallax with respect to the latest keyframe, defined as the ratio between the frame translation $\mathbf{t}$ and the mean inverse depth of the tracking local map $\bar{\rho}$: $s_t = \parallel\mathbf{t}\bar{\rho}\parallel_2$.
	% illumination change
	\item The illumination change, measured as the relative brightness transfer function between the tracked frame and the latest keyframe, i.e. $s_a = |a_{k} - a_i|$.
\end{enumerate}

A heuristic score based on the weighted combination of these criteria determines if the tracked frame is selected as a new keyframe: $w_u s_u + w_{t}s_{t} + w_{a}s_{a} > 1$.

\paragraph{New Map Point Tracking}
During exploration, the system requires to create new map points. Each keyframe contains a list of candidate points that are initialized and activated if so decided. We initialize the inverse depth of these candidate points using consecutives new tracked frames. To do so, we search along the epipolar line to find the correspondence with minimum photometric error (Eq. \ref{eq:model}). Only distinctive points with low uncertainty will be activated and inserted into the PBA.

Note that this delayed strategy requires several correspondences to obtain a good initialization as we are working with small baselines that render low parallax. To guarantee that we have enough initialized candidates to activate, we maintain candidate points from a keyframe until this is dropped from the temporal part of the LMCW. We only activate points that belong to image areas depleted from points (Sec. \ref{sec:lmcw}). Thus, when revisiting already mapped scene regions, only a few new points will be activated, as we will reuse existing map points.

\section{Results} \label{sec:results}
The proposed system is validated in the EuRoC MAV dataset \cite{Burri2015}. It has three scenarios, two rooms (V1, V2) and a machine hall (MH), with very challenging motions and changes in illumination. It also includes the 3D reconstruction ground-truth. We study the benefits of the VSLAM scheme of DSM with a version, DSM-SW (sliding-window), which uses only temporally connected keyframes as in \cite{Engel2016a}. We compare our approach against state-of-the-art algorithms such as ORB-SLAM \cite{Mur-Artal2015}, DSO \cite{Engel2016a} and LDSO \cite{Gao2018}. We evaluate the RMS Absolute Trajectory Error (ATE) and the Point to Surface Error (PSE). The ATE is computed using the keyframe trajectory for each sequence after Sim(3) alignment with the ground-truth. The PSE is estimated measuring the distance of the reconstructed model to the ground-truth surface after the trajectory alignment. The results are shown using normalized cumulative error plots, which provide the percentage of runs/points with an error below a certain threshold. These graphics provide both information about the accuracy and robustness of the evaluated method. All experiments are executed using a standard PC with an Intel Core i7-7700K CPU and 32 GB of RAM. 

\subsection{Parameter analysis and tuning} \label{sec:param_exp}
This section presents an experimental analysis of the main parameters and options defining the DSM performance. To cover more cases, we run different experiments for left and the right cameras of the stereo rig, and both in the forward and in the backward direction. We run each sequence 5 times, for a total of 220 experiments.

\subsubsection{Coarse-to-fine PBA}

\begin{figure}
	\centering
	\includegraphics[width=0.48\textwidth]{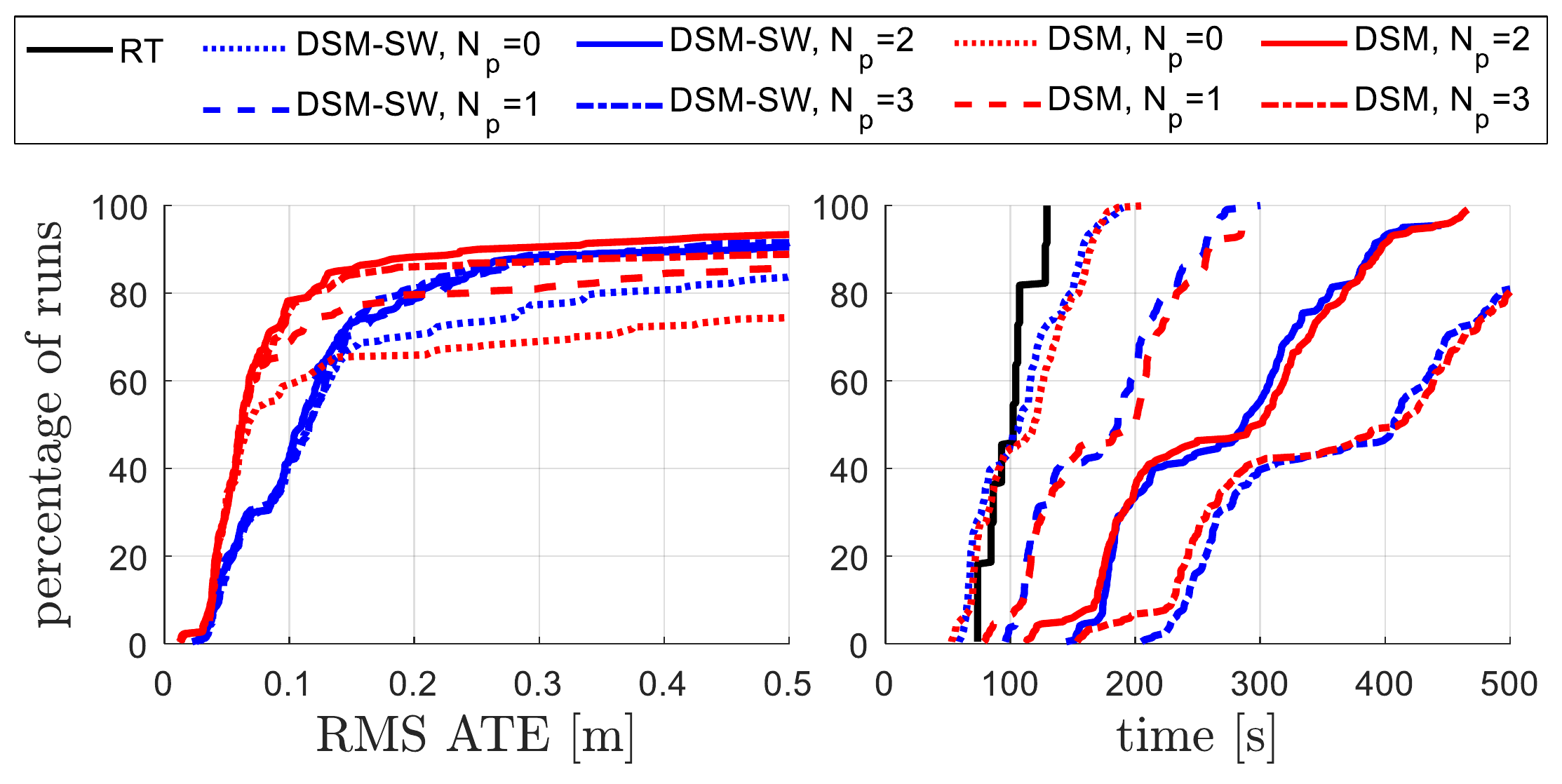}
	\caption{Number of pyramid levels $N_p$. RMS ATE (left) and processing times (right) compared with the RT (real-time) for different $N_p$.}
	\label{fig:pyramids}	
\end{figure} 

We evaluate the effect of changing the number of pyramid levels $N_p$ during the PBA. Fig. \ref{fig:pyramids} shows the results for DSM-SW and DSM. Without the coarse-to-fine scheme DSM-SW performs better than DSM. Here, DSM is not able to benefit from point reobservations due to the accumulated drift. However, DSM is able to reuse map points for higher number of pyramid levels and it clearly achieves better accuracy. While a coarse-to-fine strategy certainly increases the accuracy of DSM, there is significantly less improvement for DSM-SW. This is the expected behavior since DSM requires larger convergence radius to process reobservations while DSM-SW does not. Note how DSM is able to process approximately the 80\% of runs with a RMS ATE bellow 0.1m while DSM-SW only gets 40\% of runs. Moreover, we see that using $N_p=1$ with a sliding-window increases the performance. We also observe that increasing the number of levels after $N_p=2$ for DSM does not increase accuracy but increases the runtime significantly. 

Including reobservations in the PBA has little effect on the processing time. In contrast, the number of pyramids approximately increases the runtime by 50\% for each level. Thus, we use $N_p = 2$ as default which achieves the best balance between efficiency and accuracy.

\subsubsection{Robust Influence Function}

\begin{figure}
	\centering
	\includegraphics[width=0.38\textwidth]{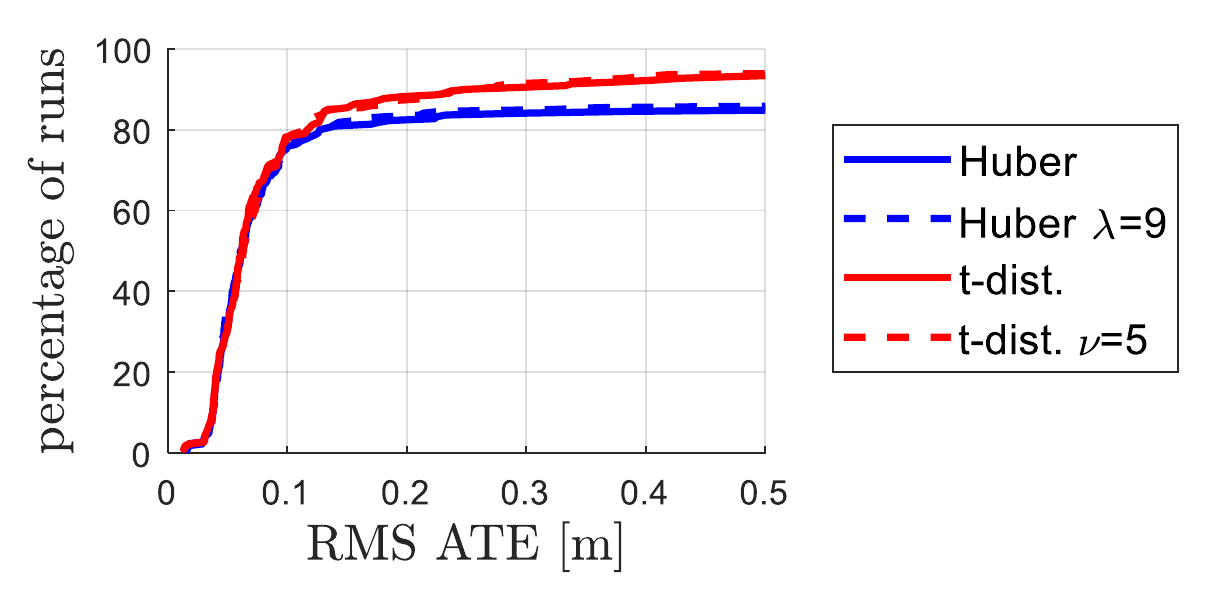}
	\caption{Robust influence function. Comparison of the RMS ATE between a Gaussian based M-estimator (Huber) and the t-distribution.}
	\label{fig:distribution}
\end{figure} 

We study the effect of the selected model of weight distribution. Fig. \ref{fig:distribution} shows the results for the t-distrution and Huber models. In contrast to \cite{Kerl2013}, we evaluate the influence of the model when the degrees of freedom $\nu$ are estimated together with the scale $\sigma$. For Huber, we study when the constant is fixed to $\lambda=9$ and when it is dynamically changed with the MAD value. Interestingly, there is not significant difference between using fixed or dynamic values on both distribution models. However, the t-distribution performs better in challenging situations providing higher robustness than Huber. This comes from the fact that the t-distribution quickly drops the weights as errors move to the tails while the Huber model does not. We use the complete t-distribution model as default settings due to its flexibility handling challenging situations.

\subsubsection{Number of covisible keyframes in the LMCW}

\begin{figure}
	\centering
	\includegraphics[width=0.38\textwidth]{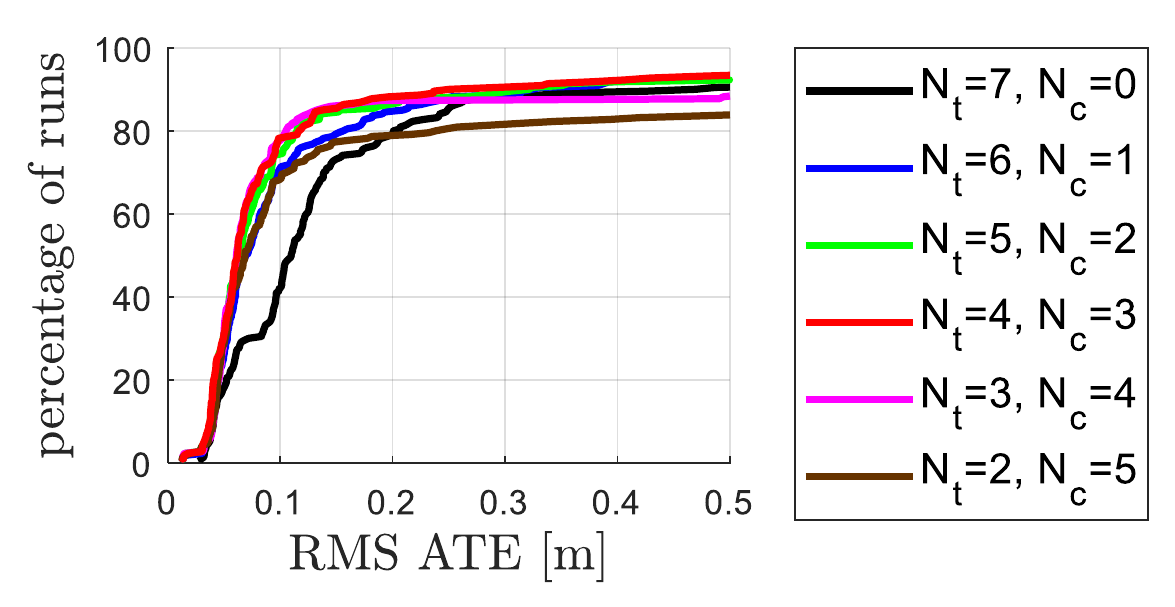}
	\caption{LMCW $N_w = N_t + N_c$. RMS ATE when changing the number of temporal $N_t$ and covisible $N_c$ keyframes.}
	\label{fig:LMCW}
\end{figure} 

We observe that increasing the number of covisible keyframes $N_c$ increases the trajectory accuracy (Fig. \ref{fig:LMCW}.) With those covisible keyframes the PBA is able to handle point reobservations and to reduce the drift. However, the system requires temporally connected keyframes $N_t$ to guarantee the odometry robustness. Taking few temporal keyframes drastically reduces the accuracy. This is due to the fact that the temporal part ensures that new keyframes are well optimized and that enough new points are initialized during exploration. Thus, we use the combination of $N_t=4$ and $N_c=3$ as default settings, which achieves the best balance between precision and robustness.
 
\subsection{Quantitative results} \label{sec:quant_exp}
This section presents a comparison of DSM against ORB-SLAM \cite{Mur-Artal2015}, DSO \cite{Engel2016a} and LDSO \cite{Gao2018}. We report the results published in \cite{Mur-Artal2016a} for ORB-SLAM, in \cite{Engel2016a} for DSO and we use the open-source implementation for LDSO. All results are obtained using a sequential implementation without enforcing real-time operation using $N_w=7$ active keyframes for all direct methods. We run on default settings all sequences both forward and backward, 10 times each, using left and right videos separately for a total of 440 runs.

\subsubsection{Trajectory error}

\begin{table}[t]
    \caption{RMS ATE [m] using forward videos for left (l) and right (r) sequences. ($\times$) means failure and (-) no available data.}
    \label{tab:ATE}
    \centering
    % @{} to remove spaces
    \begin{tabular}{@{}L{0.85cm}C{0.9cm}C{0.9cm}C{0.9cm}C{0.9cm}C{0.9cm}C{0.9cm}@{}}\hline\hline
        
        Seq. & ORB-SLAM \cite{Mur-Artal2015} & DSO \cite{Engel2016a} & LDSO \cite{Gao2018} & DSM-SW & DSM & DSM (Global PBA) \\\hline
        MH1\_l & 0.070 & 0.046 & 0.053 & 0.054 & \textbf{0.039} & 0.042 \\
        MH2\_l & 0.066 & 0.046 & 0.062 & 0.041 & \textbf{0.036} & 0.035 \\
        MH3\_l & 0.071 & 0.172 & 0.114 & 0.123 & \textbf{0.055} & 0.040 \\
        MH4\_l & 0.081 & 3.810 & 0.152 & 0.179 & \textbf{0.057} & 0.055 \\
        MH5\_l & \textbf{0.060} & 0.110 & 0.085 & 0.139 & 0.067 & 0.054 \\
        V11\_l & \textbf{0.015} & 0.089 & 0.099 & 0.099 & 0.095 & 0.092 \\
        V12\_l & \textbf{0.020} & 0.107 & 0.087 & 0.124 & 0.059 & 0.060 \\
        V13\_l & $\times$ & 0.903 & 0.536 & 0.888 & \textbf{0.076} & 0.068 \\
        V21\_l & \textbf{0.015} & 0.044 & 0.066 & 0.061 & 0.056 & 0.060 \\
        V22\_l & \textbf{0.017} & 0.132 & 0.078 & 0.123 & 0.057 & 0.053 \\
        V23\_l & $\times$ & 1.152 & $\times$ & 1.081 & \textbf{0.784} & 0.681\\    
        \hline
        MH1\_r & - & \textbf{0.037} & 0.050 & 0.054 & 0.045 & 0.039 \\
        MH2\_r & - & 0.041 & 0.051 & 0.039 & \textbf{0.039} & 0.034 \\
        MH3\_r & - & 0.159 & 0.095 & 0.187 & \textbf{0.048} & 0.035 \\
        MH4\_r & - & 3.045 & 0.129 & 0.188 & \textbf{0.058} & 0.052 \\
        MH5\_r & - & 0.092 & 0.087 & 0.131 & \textbf{0.064} & 0.052 \\
        V11\_r & - & 0.047 & 0.662 & 0.031 & \textbf{0.014} & 0.012 \\
        V12\_r & - & 0.080 & 0.208 & 0.118 & \textbf{0.046} & 0.043 \\
        V13\_r & - & 1.270 & 0.642 & 1.313 & \textbf{0.045} & 0.037 \\
        V21\_r & - & \textbf{0.027} & 0.040 & 0.032 & 0.034 & 0.030 \\
        V22\_r & - & 0.059 & 0.068 & 0.314 & \textbf{0.057} & 0.052 \\
        V23\_r & - & 0.540 & \textbf{0.171} & 0.889 & 0.528 & 0.482 \\
        \hline\hline
        
    \end{tabular}    
\end{table}

Table \ref{tab:ATE} reports the median errors for each forward sequence. Overall, we see that DSM-SW performs similarly to DSO. This is expected since both methods are based on the same sliding-window approach without a multiscale PBA. However, DSM-SW successfully executes all MH sequence, while DSO fails in MH\_03\_medium. This is probably due to the use of a more robust influence function in DSM-SW. DSM achieves higher accuracy in almost all sequences compared to the rest of direct approaches, DSO, LDSO and DSM-SW. DSO and LDSO only achieve slightly higher accuracy in a few sequences. ORB-SLAM obtains better results in V1 and V2, but DSM achieves the best performance for the MH sequences. Note that in contrast to ORB-SLAM, we do not incorporate any place recognition, pose-graph or relocalization modules. This shows the high precision of DSM is due to point reobservations and proves that DSM can achieve with only 7 keyframes comparable results to ORB-SLAM that uses tens of cameras in the local BA. In the sequence V1\_03\_difficult, DSM achieves an RMS ATE of only 7.6cm, which is by far the best performance among all the approaches tested. This sequence contains very rapid motions and illumination changes, which demonstrates the robustness of the proposed method. Besides, we successfully manage to complete all sequences and obtain an RMS ATE bellow 0.1m for all of them, except V2\_03\_difficult, where all of the compared approaches fail.

In addition, we have also evaluated the improvement due to a final global PBA at the end of each sequence. We have observed that the global PBA converges in a few iterations and only improves slightly the RMSE ATE, but with a significant increase in the computational cost. For instance, in the sequence V2\_02\_medium the global PBA optimizes fifty times more parameters with a processing time two orders of magnitude higher. The accuracy of the proposed direct local mapping scheme is very close to the result of a global PBA, but at a small fraction of the computational cost.

\subsubsection{Mapping vs Pose-Graph}

\begin{figure}
    \centering
    \includegraphics[width=0.37\textwidth]{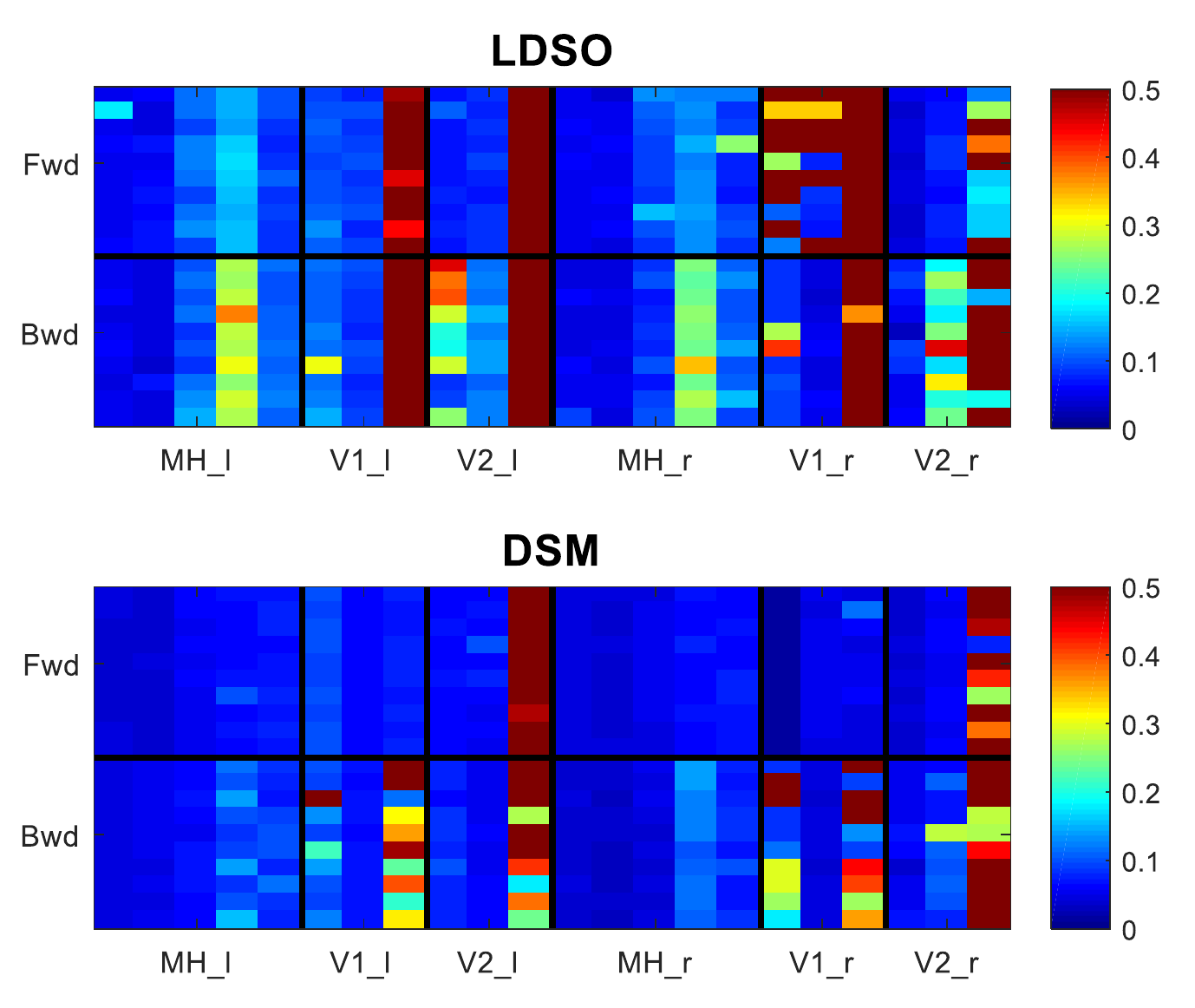}
    \caption{Full evaluation results. For each sequence (X-axis) we plot the RMS ATE [m] in each iteration (Y-axis), with a total of 440 runs.}
    \label{fig:colormap}
\end{figure} 

\begin{figure}
    \centering
    \includegraphics[width=0.37\textwidth]{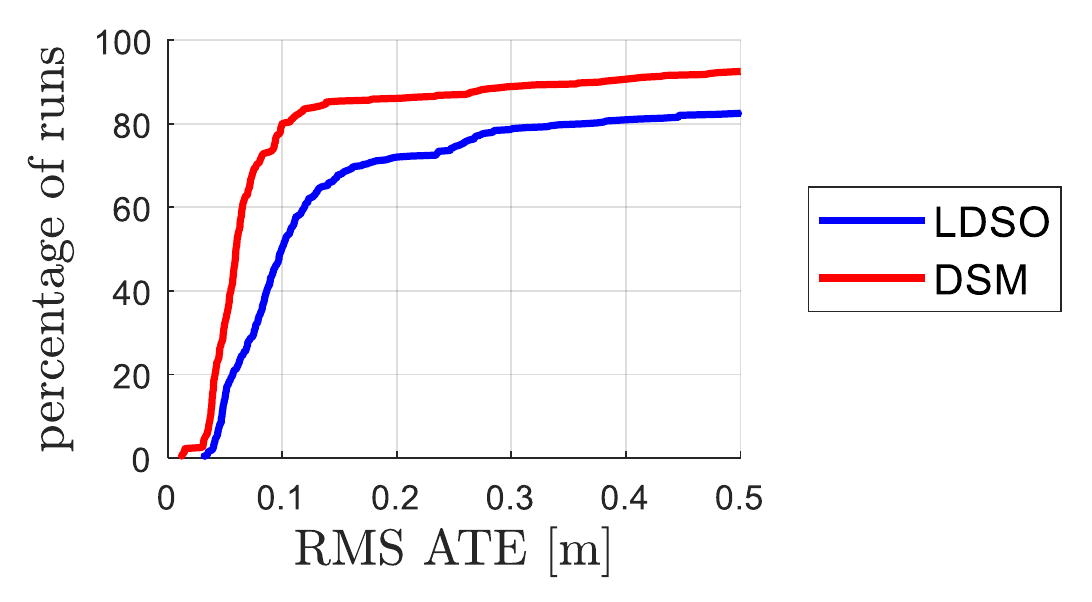}
    \caption{RMS ATE for LDSO and DSM.}
    \label{fig:ate_accum}
\end{figure} 

Comparing LDSO and DSM shows the differences in using a VO scheme with a pose-graph in contrast to a VSLAM scheme. Fig. \ref{fig:colormap} shows the RMS ATE for all the evaluated sequences for LDSO and DSM. Overall, we observe that DSM achieves better accuracy. We also see that reusing existing map points allows completing successfully a higher percentage of sequences. We build a persistent map and reuse map points to support the odometry estimation instead of permanently marginalizing all points that leave the local window. This can also be observed in Fig. \ref{fig:ate_accum}. While DSM is able to process 80\% of sequences with an RMS ATE bellow 0.1m, LDSO can only handle 50\% of runs under this limit.

\begin{figure}
    \centering
    \includegraphics[width=0.4\textwidth]{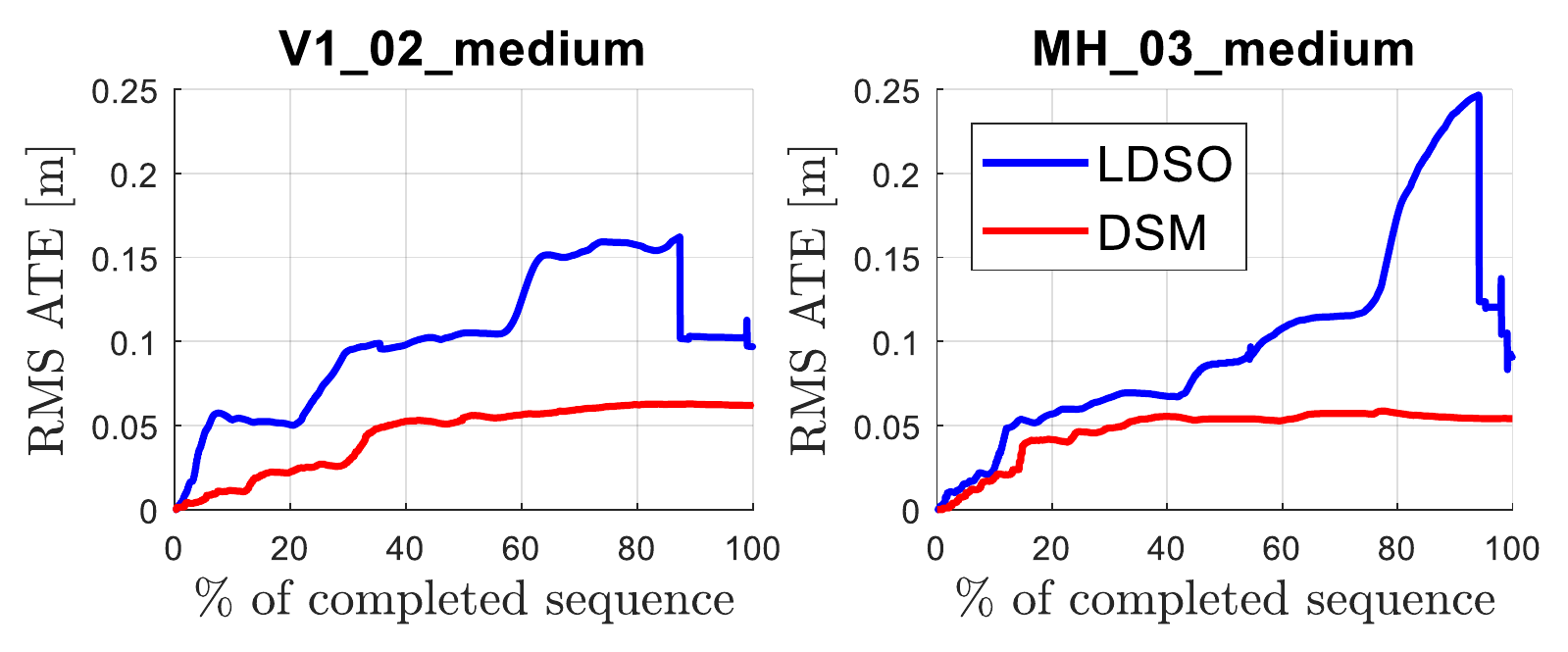}
    \caption{VSLAM vs VO + Pose-Graph. RMS ATE after processing each keyframe in the trayectory. It shows the time evolution of the error. While a feature-based pose-graph strategy may miss many loop closures, a VSLAM scheme continuously reuses existing information to provide more accurate and reliable estimates in time.}
    \label{fig:MapVSGraph}
\end{figure}

Moreover, we have observed that in some sequences LDSO misses many available loop closures due to lack of feature matches. This makes the odometry drift until a larger correction loop is detected, causing a temporally inconsistent trajectory and structure estimations. Fig. \ref{fig:MapVSGraph} shows the evolution of the RMS ATE along the trajectory. It can be seen the effect of missing loop closures with a feature-based pose-graph strategy. In contrast, building a persistent map enables reusing existing map information continuously, which maintains the trajectory accuracy stable in time. Although the final RMS ATE is similar in both systems, the odometry using a VSLAM approach is more accurate and, thus, more reliable. This clearly shows that using a VSLAM scheme provides better accuracy performance compared to a VO scheme with a pose-graph.

\subsubsection{Map error}
\begin{figure}
    \centering
    \includegraphics[width=0.43\textwidth]{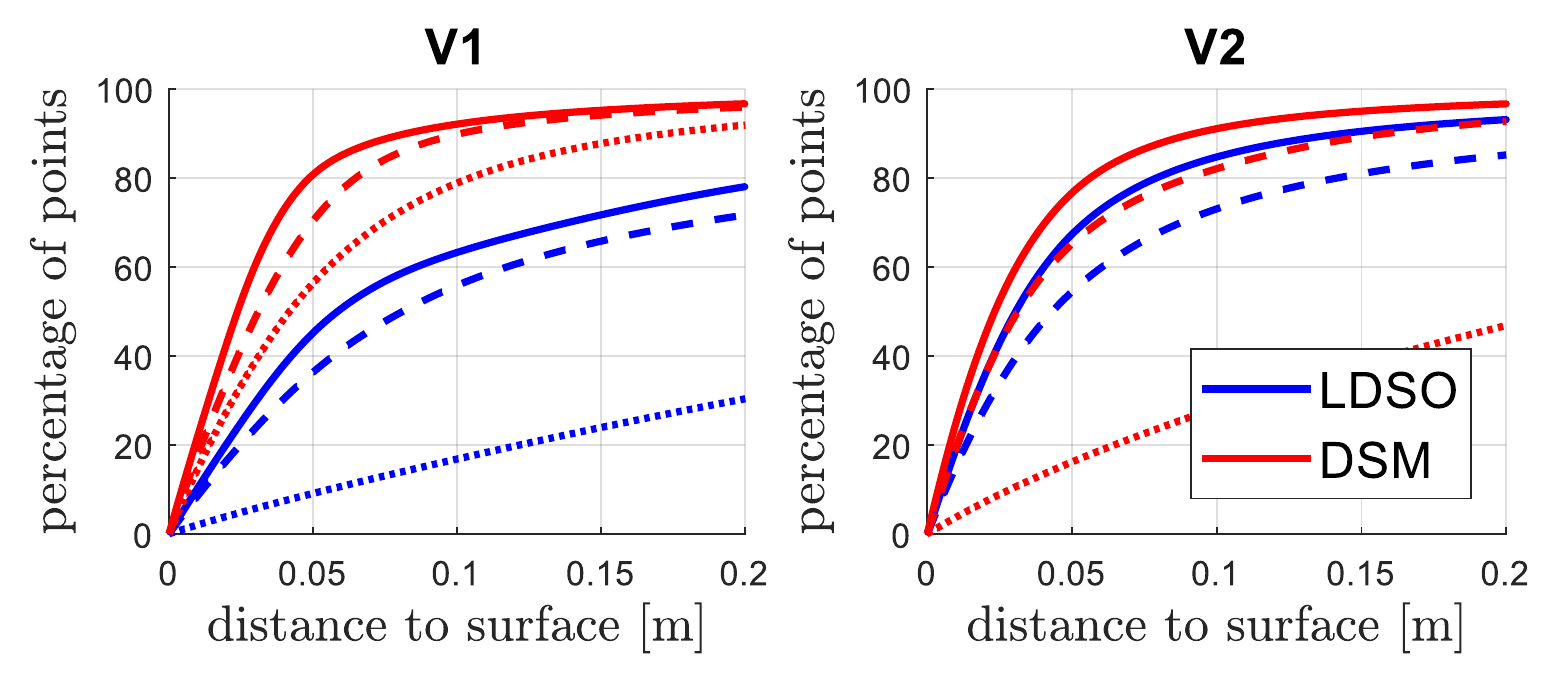}
    \caption{Map error. For each scene we show the accumulated PSE distribution using all the reconstructed 3D points for all runs. Solid lines (---) present easy sequences, dashed lines (-{}-{}-) medium and dotted lines ($\cdot\cdot\cdot$) difficult ones for each scene.}
    \label{fig:StructError}
\end{figure}

Fig. \ref{fig:StructError} shows the distance between the reconstructed points and the ground-truth surface. We compare all the sequences against LDSO except in V2\_03\_difficult where LDSO fails. Clearly, incorporating map point reobservations into the PBA increases not only the trajectory accuracy but the reconstruction precision too. Although the final trajectory RMS ATE is similar in some sequences, such as in V1\_01\_easy, the map is without a doubt more accurate in DSM. Besides, we have observed that LDSO creates ten times more points than DSM for these sequences, due to the fact that DSM reuses existing map points avoiding duplications.

\subsubsection{Processing time}

\begin{table}[t]
    \caption{Processing time and keyframe frecuency.}
    \label{tab:Time}
    \centering
    \begin{tabular}{c|rrr}\hline\hline
        
        Operation & Median [ms] & Mean [ms] & St.D. [ms] \\\hline
        Frame \& Point Tracking & 7.44 & 7.45 & 0.31 \\
        Local PBA & 888.77 & 908.53 & 121.10 \\\hline
        Keyframe Period & 396.28 & 397.22 & 177.51 \\
        \hline\hline
        
    \end{tabular}    
\end{table}

Table \ref{tab:Time} reports the processing time required for each part of the method, as well as the used keyframe period time. In our current initial implementation, PBA is the bottleneck of the processing cost. We observe that it should be twice faster to obtain the required keyframe creation rate. It is possible to improve the runtime significantly using SIMD instructions to process each patch. Besides, many of the operations can be parallelized as they are independent for each point. We believe using these upgrades could make DSM run in real-time applications since the mapping thread is not required to run at frame rate but at keyframe rate. 

\subsection{Qualitative results} \label{sec:qual_exp}
\begin{figure*}
    \centering
    \includegraphics[width=0.78\textwidth]{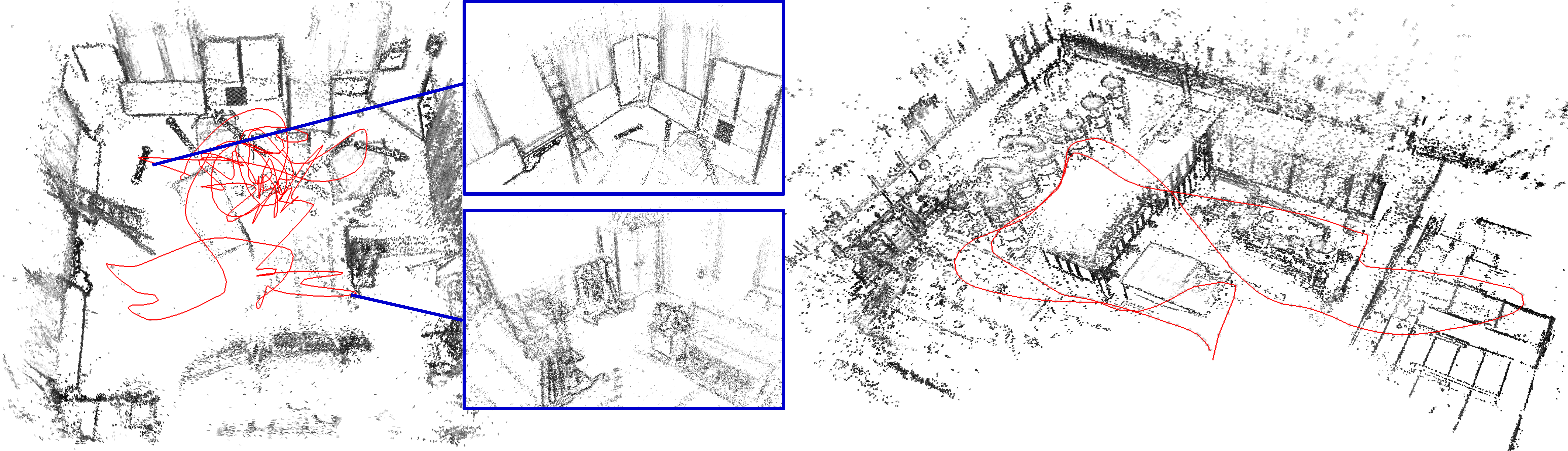}
    \caption{Qualitative examples. V1\_03\_difficult (left) and MH\_05\_difficult (right) sequences. The trajectory is displayed in red.}
    \label{fig:qual_example}
\end{figure*}

Fig. \ref{fig:example_V21} and Fig. \ref{fig:qual_example} show some 3D maps obtained with DSM. In contrast to sliding-window based approaches incorporating covisibility constraints avoid duplicating points and builds a consistent map. DSM estimates a precise camera trajectory and 3D reconstruction even in the most difficult sequences such as V1\_03\_difficult and MH\_05\_difficult (see accompanying video). 
 
\section{Discussion \& Future work}
We have demonstrated the benefits of building a persistent map instead of just estimating the camera odometry with a temporary map. Both the accuracy of the trajectory and the reconstructed map improve by reusing map information in the photometric model. DSM manages to process scene reobservations and successfully completes 10 out of 11 sequences with an RMS ATE below 0.1m in the challenging EuRoC dataset without requiring any loop closure detection and correction. During long-term sequences in the same environment DSM provides reliable estimates as long as point reobservations are successfully processed. It would be interesting to add map maintenance strategies, such as removal of redundant keyframes and points, to ensure long-term operation efficiency and allow to perform a feasible global bundle adjustment as in \cite{Mur-Artal2015}. Besides, we have shown that the t-distribution fits well the sparse photometric errors, yielding a more robust PBA. However, it would be interesting to evaluate it against other alternatives such as the Cauchy M-estimator.

Even with a persistent map, it is not possible to handle all reobservations in all situations. In large trajectory scenarios, the accumulated drift makes it impossible to detect map point reobservations with geometric techniques alone. Sometimes map point reobservations do not even fall in the camera field of view due to the large drift, e.g. in a highway loop. In these cases, a place recognition module, which exploits the image appearance, would be useful to detect loop closures. Then, a pose-graph optimization will serve as an initialization for the PBA. Therefore, we believe that combining map reuse capabilities with a place recognition module, such as previously done with indirect techniques in \cite{Strasdat2011,Mur-Artal2015}, is the best alternative. In any case, we think that a pose-graph should only be used as a coarse initialization technique for the PBA, which is the optimization technique that actually exploits all the available geometric information in a VSLAM system.

\section{Conclusion}
In this work, we have presented a novel fully direct VSLAM method which is capable of building a persistent map by reusing map points from already visited scene regions. To obtain this, we have presented a new local window selection strategy using covisibility criteria, which enables to include map point reobservations into the PBA. We have demonstrated that a coarse-to-fine strategy is required to process point reobservations with the photometric model. In addition, we have incorporated a robust influence function based on the t-distribution which increases the robustness of the whole system against spurious observation. As a result, we use the same objective function and map points for all the operations in the system. We demonstrate in the EuRoC MAV dataset that the proposed method reduces both the estimated trajectory and map error while avoiding inconsistent map point duplications at the same time. 

\ifCLASSOPTIONcaptionsoff
  \newpage
\fi

\section*{Acknowledgments}
We would like to express our gratitude to Prof. J.D. Tard\'os for the fruitful discussions and sensible advice.



\bibliographystyle{IEEEtran}
\bibliography{IEEEabrv,./references/MyCollection}

\begin{thebibliography}{10}
\providecommand{\url}[1]{#1}
\csname url@samestyle\endcsname
\providecommand{\newblock}{\relax}
\providecommand{\bibinfo}[2]{#2}
\providecommand{\BIBentrySTDinterwordspacing}{\spaceskip=0pt\relax}
\providecommand{\BIBentryALTinterwordstretchfactor}{4}
\providecommand{\BIBentryALTinterwordspacing}{\spaceskip=\fontdimen2\font plus
\BIBentryALTinterwordstretchfactor\fontdimen3\font minus
  \fontdimen4\font\relax}
\providecommand{\BIBforeignlanguage}[2]{{%
\expandafter\ifx\csname l@#1\endcsname\relax
\typeout{** WARNING: IEEEtran.bst: No hyphenation pattern has been}%
\typeout{** loaded for the language `#1'. Using the pattern for}%
\typeout{** the default language instead.}%
\else
\language=\csname l@#1\endcsname
\fi
#2}}
\providecommand{\BIBdecl}{\relax}
\BIBdecl

\bibitem{Engel2016a}
J.~Engel, V.~Koltun, and D.~Cremers, ``{Direct Sparse Odometry},'' \emph{IEEE
  Tran. Pat. Anal. and Mach. Intell.}, vol.~40, no.~3, pp. 611 -- 625, 2016.

\bibitem{Burri2015}
M.~Burri, J.~Nikolic, P.~Gohl, T.~Schneider, J.~Rehder, S.~Omari, M.~Achtelik,
  and R.~Siegwart, ``{The EuRoC MAV Datasets},'' \emph{Int. J. of Robotics
  Research}, vol.~35, no.~10, pp. 1157--1163, 2015.

\bibitem{Davison2007}
A.~J. Davison, I.~D. Reid, N.~D. Molton, and O.~Stasse, ``{MonoSLAM: Real-Time
  Single Camera SLAM},'' \emph{IEEE Transactions on Pattern Analysis and
  Machine Intelligence}, vol.~29, no.~6, p.~16, 2007.

\bibitem{Civera2008}
J.~Civera, A.~J. Davison, and J.~M.~M. Montiel, ``{Inverse depth
  parametrization for monocular SLAM},'' \emph{IEEE Transactions on Robotics},
  vol.~24, no.~5, pp. 932--945, 2008.

\bibitem{Klein2007}
G.~Klein and D.~Murray, ``{Parallel Tracking and Mapping for Small AR
  Workspaces},'' in \emph{IEEE Int. Symp. on Mixed and Aug. Reality}, 2007.

\bibitem{Strasdat2011}
H.~Strasdat, A.~J. Davison, J.~M.~M. Montiel, and K.~Konolige, ``{Double window
  optimisation for constant time visual SLAM},'' in \emph{Proceedings of the
  IEEE International Conference on Computer Vision}, 2011.

\bibitem{Mur-Artal2015}
R.~Mur-Artal, J.~M.~M. Montiel, and J.~D. Tardos, ``{ORB-SLAM: A Versatile and
  Accurate Monocular SLAM System},'' \emph{IEEE Transactions on Robotics},
  vol.~31, no.~5, pp. 1147--1163, 2015.

\bibitem{Forster2014}
C.~Forster, M.~Pizzoli, and D.~Scaramuzza, ``{SVO: Fast semi-direct monocular
  visual odometry},'' \emph{Proceedings - IEEE International Conference on
  Robotics and Automation}, pp. 15--22, 2014.

\bibitem{Leutenegger2015}
S.~Leutenegger, S.~Lynen, M.~Bosse, R.~Siegwart, and P.~Furgale,
  ``{Keyframe-based visual-inertial odometry using nonlinear optimization},''
  \emph{International Journal of Robotics Research}, vol.~34, no.~3, 2015.

\bibitem{Galvez-Lopez2012}
D.~Galvez-L{\'{o}}pez and J.~D. Tardos, ``{Bags of Binary Words for Fast Place
  Recognition in Image Sequences},'' \emph{IEEE Transactions on Robotics},
  vol.~28, no.~5, pp. 1188--1197, oct 2012.

\bibitem{Qin2018}
T.~Qin, P.~Li, and S.~Shen, ``{VINS-Mono: A Robust and Versatile Monocular
  Visual-Inertial State Estimator},'' \emph{IEEE Transactions on Robotics},
  vol.~34, no.~4, pp. 1004 -- 1020, 2018.

\bibitem{Engel2014}
J.~Engel, T.~Sch, and D.~Cremers, ``{LSD-SLAM: Large-Scale Direct Monocular
  SLAM},'' in \emph{Europ. Conf. Comp. Vis.}, 2014, pp. 834--849.

\bibitem{Cummins2008}
M.~Cummins and P.~Newman, ``{FAB-MAP: Probabilistic localization and mapping in
  the space of appearance},'' \emph{International Journal of Robotics
  Research}, vol.~27, no.~6, pp. 647--665, 2008.

\bibitem{Gao2018}
X.~Gao, R.~Wang, N.~Demmel, and D.~Cremers, ``{LDSO: Direct Sparse Odometry
  with Loop Closure},'' in \emph{IEEE/RSJ International Conference on
  Intelligent Robots and Systems}, 2018.

\bibitem{Kerl2013}
C.~Kerl, J.~Sturm, and D.~Cremers, ``{Robust odometry estimation for RGB-D
  cameras},'' in \emph{Proceedings - IEEE International Conference on Robotics
  and Automation}, 2013, pp. 3748--3754.

\bibitem{Lange1989}
K.~L. Lange, R.~J. Little, and J.~M. Taylor, ``{Robust statistical modeling
  using the t distribution},'' \emph{Journal of the American Statistical
  Association}, vol.~84, no. 408, pp. 881--896, 1989.

\bibitem{Babu2016}
B.~W. Babu, S.~Kim, Z.~Yan, and L.~Ren, ``{$\sigma$-DVO: Sensor Noise Model
  Meets Dense Visual Odometry},'' in \emph{IEEE International Symposium on
  Mixed and Augmented Reality}, 2016.

\bibitem{Agarwal}
S.~Agarwal, K.~Mierle, and Others, ``{Ceres Solver},''
  \emph{http://ceres-solver.org}.

\bibitem{Triggs2000}
B.~Triggs, P.~F.~P. McLauchlan, R.~I. Hartley, and A.~W.~A. Fitzgibbon,
  ``{Bundle Adjustment. A Modern Synthesis},'' in \emph{International Worshop
  on Vision Algorithms}, vol. 1883, 1999, pp. 298--372.

\bibitem{LagariasJeffrey1998}
C.~{Lagarias, Jeffrey}, A.~{Reeds, James}, H.~{Wright, Margaret}, and
  E.~{Wright, Paul}, ``{Convergence properties of the nelder mead simplex
  method in low dimensions},'' \emph{SIAM J. on Optim.}, vol.~9, no.~1, pp.
  112--147, 1998.

\bibitem{Baker2004}
S.~Baker and I.~Matthews, ``{Lucas-Kanade 20 years on: A unifying framework},''
  \emph{Int. J. of Comp. Vis.}, vol.~56, no.~3, pp. 221--255, 2004.

\bibitem{Mur-Artal2016a}
R.~Mur-Artal and J.~D. Tardos, ``{Visual-Inertial Monocular SLAM with Map
  Reuse},'' \emph{IEEE Robotics and Automation Letters}, vol.~2, no.~2, pp. 796
  -- 803, 2016.

\end{thebibliography}

\end{document}